\documentclass[lettersize,journal]{IEEEtran}
\usepackage{amsmath,amsfonts}
\usepackage{algorithmic}
\usepackage{algorithm}
\usepackage{array}
\usepackage[caption=false,font=normalsize,labelfont=sf,textfont=sf]{subfig}
\usepackage{textcomp}
\usepackage{stfloats}
\usepackage{url}
\usepackage{verbatim}
\usepackage{graphicx}
\usepackage{cite}
\usepackage{booktabs, multirow} 
\usepackage[hidelinks]{hyperref}

\begin{document}

\title{A Decomposition of Interaction Force for Multi-Agent Co-Manipulation}

\author{Kody~B.~Shaw, Dallin~L.~Cordon, Marc~D.~Killpack, and John~L.~Salmon}

\maketitle

\begin{abstract}
Multi-agent human-robot co-manipulation is a poorly understood process with many inputs that potentially affect agent behavior. This paper explores one such input known as interaction force. Interaction force is potentially a primary component in communication that occurs during co-manipulation. There are, however, many different perspectives and definitions of interaction force in the literature. Therefore, a decomposition of interaction force is proposed that provides a consistent way of ascertaining the state of an agent relative to the group for multi-agent co-manipulation. This proposed method extends a current definition from one to four degrees of freedom, does not rely on a predefined object path, and is independent of the number of agents acting on the system and their locations and input wrenches (forces and torques). In addition, all of the necessary measures can be obtained by a self-contained robotic system, allowing for a more flexible and adaptive approach for future co-manipulation robot controllers. 

\end{abstract}


\IEEEpeerreviewmaketitle

\section{Introduction}

\IEEEPARstart{H}{UMANS} thrive by working together to complete tasks that could not be accomplished individually; they experience efficiency gains through group efforts. Developing a sound understanding of the way people coordinate their efforts and work together is an essential step in creating a robot that can assist, or even replace human agents in cooperative tasks \cite{takagi_individuals_2019}. Perhaps one of the most basic forms of such cooperation is co-manipulation, which we define as collaboration by two or more individuals to move a significant load or extended object. Examples requiring two or more agents might include moving furniture in a home, carrying materials at a construction site, or supporting a stretcher in a search-and-rescue situation.

Co-manipulation tasks are often analyzed from the perspective of interaction force, which is one component of haptic signals that occur during co-manipulation. Although researchers seem to agree that interaction force is important, there is disagreement on how to clearly define this haptic feature and its purpose. Some groups use the term to refer to any forces generated by human partners involved in an interaction \cite{reed_kinesthetic_2005, bussy_proactive_2012, erhart_internal_2015, schmidts_new_2016, sawers_small_2017, lanini_interactive_2017}. Other researchers define the interaction force to be the component of human applied forces that result in no net work or acceleration of the object thus not contributing to the magnitude of the net force, as might be the case when two individuals push equally against one another \cite{kumar_force_1988, groten_experimental_2009, mortl_role_2012, noohi_model_2016, Mielke_HR_data_driven_2024, jensen_trends_2021}. 

As for the role of the interaction force, there has been a great deal of discussion. Jlassi et al. \cite{jlassi_online_2014} suggest that interaction forces signal a change, and lowering the amplitude of the interaction force decreases comfort. Aligning with the first definition and describing force control, Cremer et al. \cite{cremer_robotic_2014} state that, ``the objective is to maintain a desired interaction force."  Noohi \cite{noohi_model_2016} claims that the interaction force contains critical information for improving performance. The three examples here are just a small sample of the many papers discussing interaction forces \cite{bussy_proactive_2012, mortl_role_2012, sawers_small_2017, lanini_interactive_2017, Mielke_HR_data_driven_2024, takagi_individuals_2019,  schmidts_new_2016, donner_physically_2018, erhart_internal_2015}. Evidently, interaction force is widely considered a key component in understanding the intuitive act of physical co-manipulation. In this paper, we will be discussing interaction force as defined by the measured forces at the agent-object boundary.

Our research focuses on the development of robots designed to assist in bearing loads requiring the collaboration of two or more individuals. These tasks typically consist of maneuvering an object of significant mass or size through real-world environments, necessitating effective haptic communication. Interaction force is among the most common measures in works of a similar type. However, in pursuing these objectives, inherent challenges and key gaps in the existing definitions and models of interaction force present us with decisions that, to the best of our knowledge, have not been fully addressed. Our goal is to determine a decomposition of interaction force in light of the following requirements: 
\begin{itemize}
    \item relies only on the sensors available on a self-contained robot
    \item is independent of the choice of frame
    \item extends to six degrees of freedom (DOFs)
    \item handles objects of significant mass
    \item is independent of the number of agents
\end{itemize}

This paper defines a decomposition of interaction force that meets these requirements and explores insights into data from two different human-human (HH) co-manipulation studies and one human-robot study using this measure.

\subsection{Prior work}
\IEEEpubidadjcol

As presented in the introduction, there is a wide range of discussion regarding interaction force, which, unsurprisingly, includes several contradictory ideas. For example, Bussy et al. \cite{bussy_proactive_2012, mortl_role_2012} claim that interaction force is a wasted force that should be minimized. Meanwhile, Sawers et al. \cite{sawers_small_2017} found that experts exert higher interaction forces than novices, suggesting that increased interaction forces improve communication. In other studies, researchers hypothesized that interaction force encourages stability and communicates intent \cite{lanini_interactive_2017, Mielke_HR_data_driven_2024, takagi_individuals_2019}. Initially, minimizing this interaction force seems to be the obvious solution, \cite{karayiannidis_online_2013, duchaine_general_2007, ikeura_cooperative_1995}, but further investigation through (HH) studies suggests other objectives that do not seek the minimization of this force or effort \cite{Mielke_HR_data_driven_2024, jensen_trends_2021,sawers_small_2017, jlassi_online_2014, freeman_motion_2024}. 

Along with disagreements about the purpose of interaction forces, there is a general lack of consideration regarding how interaction force should be defined. Many studies that discuss interaction force offer descriptions or adjectives instead of strict mathematical definitions, thus leaving a great deal of room for interpretation and little room for use in cross-study comparisons. To give just a few examples: Parker and Croft \cite{parker_experimental_2011} describe it as a “tensioning force”; and Corteville et al.\cite{corteville_human-inspired_2007} offer no description or definition, though they imply that it means the force applied by a user or operator. Several more examples not detailed here include \cite{corteville_human-inspired_2007, parker_experimental_2011, sawers_small_2017, lanini_interactive_2017, melendez-calderon_interpersonal_2015, takagi_motion_2016, bussy_proactive_2012}. 

Some studies propose a specific model of the cause or generation of interaction force. For example, Takagi et al. \cite{takagi_physically_2017, takagi_haptic_2018, takagi_individuals_2019, takagi_flexible_2021} define interaction force as a spring and damper system that can be tuned or modeled to improve performance in simulated tasks. Unfortunately, this definition of interaction force is typically confined to simulations and is of limited utility in physically coupled systems. Peternel et al. \cite{peternel_towards_2016} address this problem by assuming knowledge of a desired state and comparing it to the current state and the measured force of the robot. Noohi et al. \cite{noohi_model_2016} offer a more unique and complex model that determines the interaction force for reaching movements. They use prior knowledge of the starting and ending position and an assumption of minimum jerk trajectories to determine forces not directed toward the goal.

A few papers offer a mathematical definition of interaction force, including Reed et al. \cite{reed_kinesthetic_2005} who give a definition for a \textit{difference force}, then proceed to use interaction force and difference force interchangeably in their following papers \cite{reed_haptic_2006, reed_physical_2008}. This interchangeable use of ``interaction'' and ``difference" force works under the assumption that the forces from the two participants are never in the same direction. This assumption is limiting in that it does not allow for active assistance from a follower agent, a participant who is not directly informed of the goal but must interpret signals from a leader and try to assist. Groten et al. \cite{groten_experimental_2009} offer a widely used definition of interaction force, which can be summarized as the lesser of two opposing forces in a two-agent system. Several papers cite Groten et al. for their definition of interaction force, \cite{mortl_role_2012, jensen_trends_2021, freeman_motion_2024, mielke_analysis_2017, madan_recognition_2015}. Unfortunately, this definition is insufficient for tasks with more than two agents, objects of significant mass, and tasks involving multiple DOFs.  Jensen et al. \cite{jensen_trends_2021} seek to expand Groten's definition of interaction force to six DOFs by simply applying Groten's definition to each DOF. This causes an issue where the interaction force becomes dependent on the frame from which it is viewed. They also claim that interaction forces can be ambiguous under the influence of gravity and that the effect of gravity should be removed, though they offer no arguments or supporting evidence for those statements. 

Most recently in the domain of robotic grippers, Schmidts, Donner, Erhart et al.  \cite{schmidts_new_2016, donner_physically_2018, erhart_internal_2015} have begun addressing some of the shortcomings and inconsistencies of prior decompositions of interaction force. They have made a great deal of progress toward obtaining a mathematically rigorous definition of interaction force, primarily in the realm of adding constraints to ensure physical plausibility. However, due to their reliance on knowledge of the locations and measurements of all inputs to the system, their approach is insufficient for the stated goals of this paper.

\section{Developing Net Force Decomposition of Interaction Force}

Our goal was to find a consistent way to decompose interaction forces and measure how people in groups of two or more interact through extended objects of significant mass in 6 DOFs. To start, we define interaction force as the force measured at the agent system boundary. In other words, this is the force that is measurable at the grasp point of the agent ($f_{self}$). This force can then be decomposed into two components, one that contributes to the motion of the object and another that is canceled out. There are, however, many different ways to perform this decomposition. For example, Erhart, Zuo et al. \cite{erhart_internal_2015},  \cite{bing-ran_zuo_equivalence_1999} calculate what they refer to as internal forces; Bussy \cite{bussy_proactive_2012} finds wasted forces; and Mortl et al. \cite{mortl_role_2012} calculate squeeze forces or an internal wrench. Corteville et al. \cite{corteville_human-inspired_2007},  and Parker and Croft \cite{parker_experimental_2011}, calculate tensioning forces between two participants. These decompositions of interaction force are all geared towards the studies in which they were developed and do not generalize well into other studies, making comparisons difficult. Donner et. al. \cite{donner_physically_2018} point out that there may be an infinite number of different decompositions, therefore until a general solution can be found our choice of which decomposition to use should be carefully considered.

\subsection{Addressing Requirements}

While exploring the solution space for the best decomposition to meet our constraints, several key issues became clear. Each of those issues and the decisions made concerning them are discussed in detail below. 

\subsubsection{Two or More Agents}

There were many potential ways of addressing the multiple agents/inputs problem. Takagi et al. \cite{takagi_individuals_2019} modeled the interaction force of multiple agents with an ideal spring connecting each agent to the average position of the group. This approach is impractical for our purposes because it assumes knowledge of the position of each person in the group. Several other studies (see \cite{erhart_internal_2015, schmidts_new_2016, donner_physically_2018}) consider multiple input locations from the perspective of grasping, where considerable analogy can be drawn between multi-fingered grasps and multi-agent co-manipulation. 

A decision had to be made whether to base our decomposition on information that could be obtained by placing a sensor at each participating agent or to limit the available information to that which could be measured or deduced by the onboard sensors of a single robot that is part of the co-manipulation team. A decomposition of interaction force that does not rely on a sensor being present at every point of force and torque input to the system will be more practical and flexible for use in future applications and co-manipulation research.

\subsubsection{Six Degrees of Freedom}

The question of how exactly to extend past ideas to six DOFs is another important issue in this exploration. The extension from one to six DOFs was attempted in \cite{jensen_trends_2021, mielke_analysis_2017, Mielke_HR_data_driven_2024, townsend_estimating_2017}. Jensen et al. \cite{jensen_trends_2021} provide an algorithm proposing how to calculate interaction force for a 6-DOF system. However, they independently applied Groten's \cite{groten_experimental_2009} definition of interaction force to each axis, which causes it to vary with respect to the chosen frame. Schmidts and Erhart, \cite{schmidts_new_2016, erhart_internal_2015}, also propose a way to extend interaction force to six DOFs. Their approach was then extended by Donner \cite{donner_physically_2018} to meet additional physical plausibility constraints, such as restricting the interaction force to be less than or equal to the magnitude of input forces. Unfortunately, their approach is also insufficient for our purposes due to their reliance on knowledge of the grasp matrix and force-torque measurements at all end effectors. Therefore, our approach must be able to handle three-dimensional vectors and must not depend on knowledge that would be inaccessible to a single individual (human or robot agent) in a multi-agent co-manipulation scenario.

A consistent measure of the torques of a robot relative to the net torque of the group will also be necessary to generalize to all six DOFs. Several papers have addressed torque in the context of interaction force \cite{donner_physically_2018, schmidts_new_2016, erhart_internal_2015}. Unfortunately, the approaches in these papers have already been ruled out for the reasons stated above.

\subsubsection{Gravity compensation}

 Extending our definition to include objects of significant size and mass raises questions about how the force of gravity should be incorporated. Jensen et al. \cite{jensen_trends_2021} argued that when manipulating an object of significant mass the force of gravity could make the interaction force ambiguous. Unfortunately, they offered no supporting arguments or evidence for this statement, and the authors find their approach to gravity compensation inconsistent, as it is applied only occasionally and under specific conditions. Several other potential methods of gravity compensation were explored; however, to obtain a rigorous compensation of gravity, a model of the system that includes the mass, orientation, and locations of force inputs relative to the center of mass would be required. Given our condition that all information must be obtained from a self-contained robot, these measures will not generally be available. To compensate for gravity and meet our isolation constraint, one possible approach is to consider all vertical forces and their resulting torques to be due to gravity. Although this method decreases our DOFs from 6 to 4 by ignoring forces in one axis and torques about another, we are unlikely to lose key communication because of the significant size and mass of the objects considered in the data from the experiments that we present in this paper. Any vertical forces or torques produced by an agent that would be with or against gravity must be of sufficient magnitude, against the already present gravitational forces, before they could be noticed by another participant as useful communication \cite{allin_measuring_2002}. While simple, this approach is consistent and sufficient as a first step towards a better decomposition of interaction force in this paper and the existing literature as a whole.

\subsection{Interaction Force Decomposition}

\begin{figure}[!htb]
    \includegraphics[width=0.99\columnwidth]{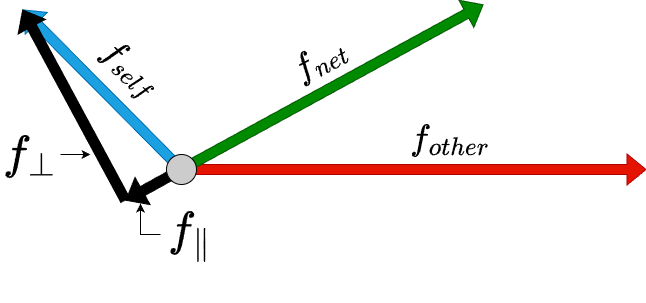}
    \caption{Net Force Method: Visualisation of $f_\parallel$ and $f_\perp$}
    \label{fig:perp_ll_to_net_force}
\end{figure}

Exploring the solution space given our objectives (obtain a decomposition of interaction force for use in human-robot co-manipulation with objects of significant size and mass, two or more agents, and 6 DOFs) and the points in the previous section yielded a few potential solutions. Among those solutions, it was determined that a method herein referred to as the \textit{net force method} had the best ratio of pros to cons. The net force method decomposes the input forces into perpendicular and parallel components to the net force. The component perpendicular to the net force ($f_\perp$) does not contribute to the magnitude of the net force but does cause a change in the direction of the net force of the object. The component of $f_{self}$ that is parallel to the net force ($f_\parallel$) either opposes or contributes to the magnitude of acceleration in the direction of the net force. Positive values of this force indicate cooperation or active contribution to the current state of acceleration. These components are pictured in Figure \ref{fig:perp_ll_to_net_force}. Understanding more about the forces in this decomposition of interaction force, and how humans behave relative to these forces, will be useful for human-robot (HR) co-manipulation because it meets the physical plausibility constraints set forth by Donner \cite{donner_physically_2018}, provides a measure that is consistent in any chosen frame, can be extended to six DOFs, allows for any number of agents, and utilizes information that is reasonably accessible to a self-contained robotic system via a force torque sensor and an accelerometer at the grasp point(s) of a robot or agent acting on a system.

The components of the net force method are calculated as follows:
\begin{equation}
\label{eq:perp}
f_\perp = \frac{f_{self}\cdot f_{net}}{||f_{net}||}\cdot \frac{f_{net}}{||f_{net}||}-f_{self}
\end{equation}

\begin{equation}
\label{eq:parallel}
f_\parallel = \frac{f_{self}\cdot f_{net}}{||f_{net}||}\cdot \frac{f_{net}}{||f_{net}||}
\end{equation}

where $f_{self}$ is the measured force between the agent or self and the system, $f_{net}$ is the calculated net force via acceleration or change in position, $f_\parallel$ and $f_\perp$ are the parallel and perpendicular components of $f_{self}$ to the net force, $\cdot$ is the dot product, and $||f_{net}||$ is the 2-norm. 

Since $f_\parallel$ and $f_\perp$ only depend on the direction of the net force and $f_{self}$ for the agent of interest, it is completely independent of the number of agents and does not depend on prior knowledge of a desired path or locations of input wrenches. These measures can also be applied to six DOFs. Thus, we can use $f_\parallel$ and $f_\perp$ as a consistent measure to inform an agent about its state relative to the state of the group.
This method can also be extended to handle torque vectors, without any alteration, with equivalents of $f_\parallel$ and $f_\perp$ to torsional forces, and thus $\tau_\parallel$ and $\tau_\perp$ can be calculated in the same manner. Having obtained this consistent measure, it should be possible to find patterns in the parallel and perpendicular force that can be mapped to appropriate behaviors in robotic control. In future work, modeling how humans respond to changes in $f_\parallel$ and $f_\perp$ over time will be instructive on how to provide active and intuitive assistance via a robot partner. This base level of control would be most helpful as a lower-level controller before we seek to add higher-level controls involving path planning and obstacle avoidance.

\section{Net Force Decomposition From Co-manipulation Data}

\begin{figure*}[htbp]
\centering
    \subfloat[Categories of Interest\label{fig:ranges}]{%
        \includegraphics[width=0.41\textwidth]{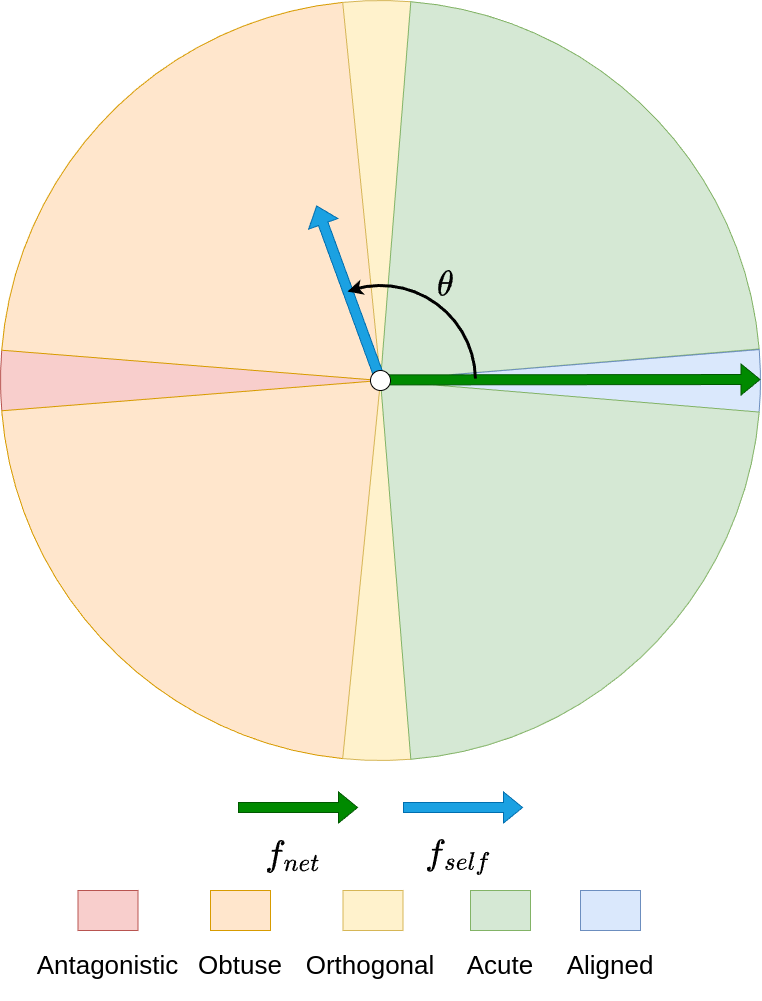}}
    \hfill
    \subfloat[Circular density plot\label{fig:cir_hist}]{%
        \raisebox{5pt}[0pt][0pt]{\includegraphics[width=0.59\textwidth]{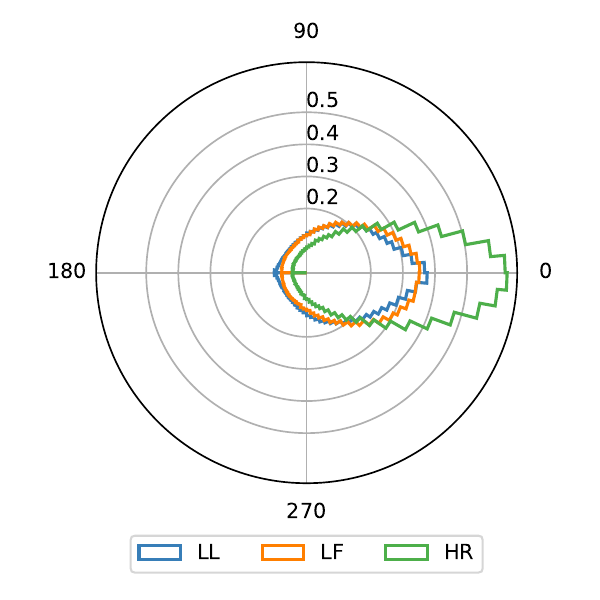}}}
    \caption{\textbf{(a)} Ranges of angles, $\theta$, between $F_{net}$ and $F_{self}$. Ranges are chosen for the categories of interest. Aligned orthogonal and antagonistic ranges are each within $\pm$ 5 degrees of 0, $\pm$90, and 180 degrees respectively, while acute and obtuse ranges cover the remaining area. \textbf{(b)} Density plot showing the relative frequency of angles $\theta$ between $F_{self}$ and $F_{net}$ across each study by 5-degree increments.}
    \label{fig:subfigures}
\end{figure*}

In an effort to obtain a sense of what this decomposition might look like in real-world scenarios and determine if it is a useful decomposition, we explored five cases of interest using data from the following three studies. First, a HH study with an assigned human leader and human follower role will be referred to as the leader-follower study (LF) \cite{mielke_analysis_2017}. Next, a human-robot (HR) study exploring different control methods on the translation vs. rotation problem between a robot and a human will be referred to as the HR study \cite{Mielke_HR_data_driven_2024}. In this HR study, it was noted that people still preferred to work with other people. For this reason, we highlight notable differences between HH and HR behaviors that may help explain that sentiment. Finally, a HH study where both human participants were given knowledge of the goal will be referred to as leader-leader (LL) \cite{freeman_motion_2024}. Each of these studies is described in greater detail in Appendix \ref{sec:appendix}. Since we explore $f_\parallel$ and $f_\perp$ in real data with non-negligible noise, specific ranges of interest were chosen to be a representative sample. These ranges are divided into five categories, three of which consist of narrow ranges of $\pm$ 5 degrees of directly aligned, orthogonal, and anti-aligned or antagonistic to the net force. These adjectives do not refer to direct interaction between agents but to comparisons with the direction of the net force. The areas between these categories are also considered and referred to as acute and obtuse with respect to the net force. Each of these five categories and their ranges are shown in Figure \ref{fig:ranges} with angles as measured by the reference frame for $\theta$. By considering the planar forces and finding the relative difference in angle from $F_{self}$ and $F_{net}$ to the x-axis of the object, we can produce a density plot showing an approximate distribution of the angle $\theta$ between $F_{self}$ and $F_{net}$ for all trials, as shown in Figure \ref{fig:cir_hist}. It is of interest to note that for both the LL and LF studies the histograms are quite similar to each other relative to the HR study. This suggests that there is more in common between the two HH studies, with dissimilar tasks, than between the similar LF and HR studies.

\begin{table}[tb]\centering
    \caption{Average magnitudes of acceleration across the five cases of interest.}
    \label{tab:mag_Acc_table}
    \begin{tabular}{lrrrr}\toprule
    &LL $\frac{m}{s^2}$ &LF $\frac{m}{s^2}$ &HR $\frac{m}{s^2}$ \\\midrule
    Aligned &0.989 &0.533 &3.760 \\
    Acute &0.845 &0.429 &2.624 \\
    Orthogonal &0.716 &0.292 &1.053 \\
    Obtuse &0.685 &0.258 &0.862 \\
    Antagonistic &0.685 &0.244 &0.764 \\
    \bottomrule
    \end{tabular}
\end{table}

We hypothesize that due to the significant mass (e.g. 20-30 kg) of the co-manipulated object used in these studies, participants will have time to actively assist in accelerating the object, rather than assume roles to start and stop the object as found by Reed et al. \cite{reed_physical_2008}. This hypothesis is supported by the data in Table \ref{tab:mag_Acc_table}, where we observe that in every study (e.g. LL, LF, HR) there is a general trend of decreasing acceleration as we move from the alignment range where $f_{self}$ is within $\pm 5$ degrees of the net force, to the antagonistic range where $f_{self}$ is 180 $\pm 5$ degrees of the net force. Averages are calculated using all data that lies within each category across each study. Note that the large differences in magnitudes of average acceleration between studies are likely due to the length of the task and noise inherent within each study.

\begin{figure}[tb]
    \includegraphics[width=0.99\columnwidth]{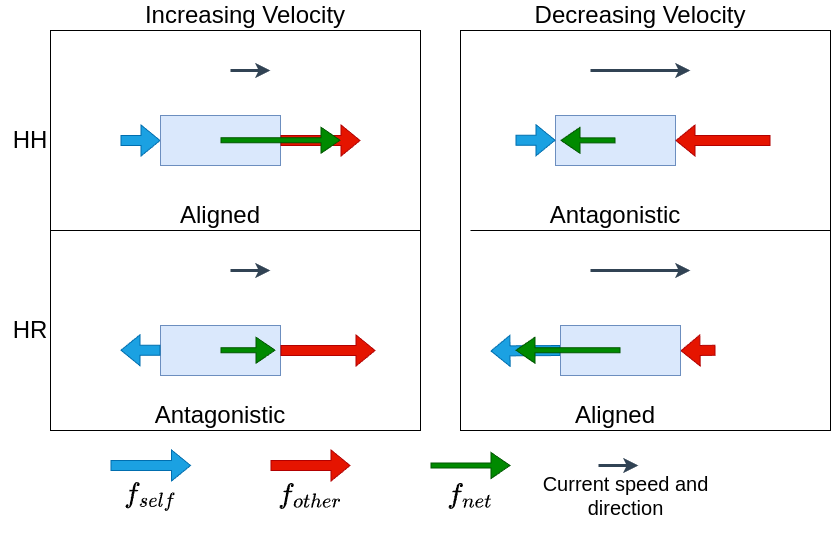}
    \caption{A comparison of HH and HR behaviors during increasing and decreasing velocity periods. Note: human followers tend to align their forces when the speed of the system is increasing while robots are antagonistic. The opposite behaviors are also observed when the system is slowing down during which humans become antagonistic while robots are aligned.}
    \label{fig:Acc_HH_HR}
\end{figure}

\begin{table}[tb]\centering
    \caption{Average signed magnitudes of acceleration across the five cases of interest. (+) is increasing the magnitude of velocity  and (-) is decreasing the magnitude of velocity}
    \label{tab:signed_Acc_table}
    \begin{tabular}{lrrrrrr}\toprule
    &Aligned &Acute &Orthogonal &Obtuse &Antagonistic \\\midrule
    LL $\frac{m}{s^2}$ &0.010 &-0.006 &-0.008 &-0.006 &-0.002 \\
    LF $\frac{m}{s^2}$ &0.370 &0.121 &-0.341 &-0.460 &-0.414 \\
    HR $\frac{m}{s^2}$ &-4.252 &-0.427 &1.139 &1.315 &1.342 \\
    \bottomrule
    \end{tabular}
\end{table}

Further insights can be gained if we consider whether the magnitudes of acceleration within each category, defined in Figure \ref{fig:ranges}, on average increase or decrease the magnitude of velocity. Positive values within a category would indicate that the range is typically used for increasing the average speed, and negative values would indicate that the range(s) are being used, on average, to come to a stop. Values near zero will tell us that the roles in increasing and decreasing velocity within the range(s) are balanced. We hypothesize that better communication will tend towards a greater balance because both participants will contribute more actively to both starting and stopping. Inspecting Table \ref{tab:signed_Acc_table}, we see, as expected, that the LL study (which we can assume to have ideal communication because both participants were given the same known goal) has the most balanced results in each category, followed by the LF study, then the HR study. A closer inspection of Table \ref{tab:signed_Acc_table} shows an interesting change in sign between the HH studies and the HR study. Note that within the aligned and antagonistic ranges, the signed average accelerations in the HR study have the opposite sign of those in the HH studies. This indicates a fundamental difference in behavior between current robot algorithms and humans. This change in sign indicates that when the leader or group is increasing the velocity of the table, human participants tend to be helpful and align their forces with the net force. This behavior reverses when the group is slowing down or coming to a stop and human followers tend towards the antagonistic range. The robots in Mielke et al. study \cite{Mielke_HR_data_driven_2024} exhibit the opposite trends in behavior - helpful when coming to a stop and tending to the antagonistic range when speeding up. This idea is demonstrated in Figure \ref{fig:Acc_HH_HR}, which shows hypothetical cases demonstrating the ideas found in Table \ref{tab:signed_Acc_table}. The following subsections explore the five categories, aligned, acute, orthogonal, obtuse, and antagonistic, in detail and compare hypothesized results to actual results.

\begin{table}[tb]\centering
\caption{Percentage time spent in the five categories of interest of $f_{self}$ with respect to the net force. Percentages are calculated by dividing the total number of data points reported by the number of data points that fall within each category.}
\label{tab:Angle_table}
\begin{tabular}{lrrrrrr}\toprule
&Aligned &Acute &Orthogonal &Obtuse &Antagonistic \\\midrule
LL &5.8\% &62.0\% &4.5\% &26.2\% &1.45\% \\
LF &6.1\% &63.9\% &4.1\% &24.6\% &1.37\% \\
HR &10.9\% &70.9\% &2.8\% &14.6\% &0.78\% \\
\bottomrule
\end{tabular}
\end{table}

\begin{figure}[tb]
    \includegraphics[width=0.99\columnwidth]{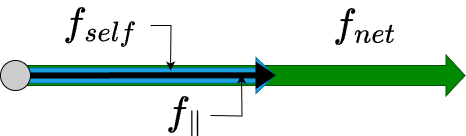}
    \caption{Alignment range occurs when the agent force and the net force are within 5 degrees}
    \label{fig:aligned}
\end{figure}

\subsection{Measured Force and Net Force are Aligned}

In the aligned case, $f_\perp$ is small relative to $f_{self}$, and $f_\parallel$ is a positive vector that is nearly equal in magnitude to $f_{self}$. It can be said that the participant is in agreement with the current direction of acceleration. However, it may not be said that they are in perfect balance because their magnitudes are not necessarily equal, as shown in Figure \ref{fig:aligned}. In this example, $f_\parallel$ contains all of the agent's force and the perpendicular component is zero. In terms of the highest efficiency, this is the ideal scenario. For our real-world analysis, we defined this scenario to occur when the force of the agent of interest is within $\pm$ 5 degrees of the net force. This situation occurs about 5.8$\%$ of the time in the LL study, 6.1$\%$ of the time in the LF study, and 10.8$\%$ of the time in the HR study (see Table \ref{tab:Angle_table}) where percentages of time are calculated by dividing the number of data points within a category by the total number across each study. This is possible because data was collected at a constant rate within each study. It is surprising to note that the smallest percentage of time spent in this state of aligned forces occurs in the LL study where both participants know the goal and should have the highest levels of cooperation. This suggests that people may prefer a state other than that of the highest efficiency. It is suspected that people are instead optimizing for comfort. This idea is discussed further in Section \ref{sec:TC} on tension versus compression.
 
\begin{figure}[tb]
    \includegraphics[width=0.99\columnwidth]{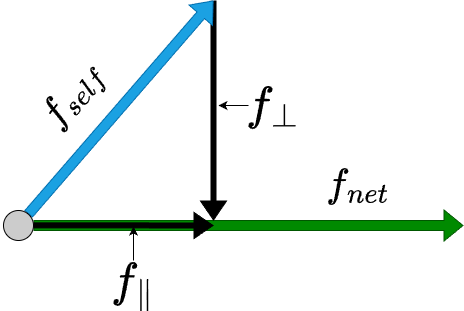}
    \caption{Acute angles with respect to the net force.}
    \label{fig:acute}
\end{figure}

\subsection{Measured Force and Net Force are Acute}

In the acute ranges of angles, $f_\parallel$ is a positive vector contributing to the magnitude of acceleration, and $f_\perp$ is a positive vector working to redirect, rotate, or stabilize as shown in Figure \ref{fig:acute}. Under the hypothesis that people aim to minimize effort and maximize efficiency and given the inherent imprecision of HH teams, we would predict that a majority, or at least a significant amount, of time would be spent in the acute ranges, specifically within $\pm $ 90 degrees of the net force and outside of the aligned range. These ranges occur 70.2$\%$ of the time in the LL study, 72.1$\%$ of the time in the LF study, and 83.2$\%$ of the time in the HR study (see table \ref{tab:Angle_table}). The LL study had the smallest percentage of time spent in this state, further supporting the notion that people prefer a state other than one that maximizes efficiency or minimizes effort. However, a large majority of time was spent in this state, regardless of the study. This implies that while it may not be the preferred state, it is probably not an undesirable or uncomfortable state.

\begin{figure}[tb]
    \includegraphics[width=0.99\columnwidth]{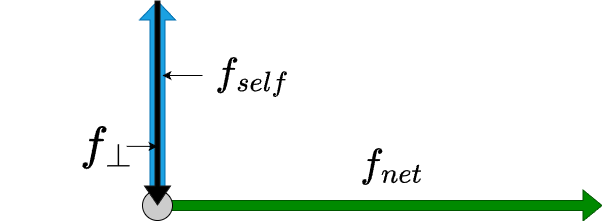}
    \caption{When $f_{self}$ is orthogonal to the net force.}
    \label{fig:orthogonal}
\end{figure}

\subsection{Measured Force is Orthogonal to the Net Force }

In this case, $f_\perp$ is reaching its max and $f_\parallel$ is small relative to $f_\perp$ as shown in Figure \ref{fig:orthogonal}. This is an interesting case where the agent is acting at 90 degrees to the net force. This perpendicular force does not contribute to the magnitude of acceleration but does contribute to the direction of the net force.

The data for this scenario was assigned when the agent's force is $\pm$ 5 degrees of the orthogonal to the net force, which occurs around 4.5$\%$ of the time in the LL study, 4.1$\%$ of the time in the LF study, and 2.8$\%$ of the time in the HR study (see Table \ref{tab:Angle_table}). Here, we begin to see the trend reverse; the LL study spent more time in this state than the other studies. Our hypothesis for this trend is that participants take on differing roles and are not always trying to contribute to the acceleration of the table but may be filling some other role, such as redirecting the table or allowing for a natural side-to-side swing caused by walking.

\begin{figure}[tb]
    \includegraphics[width=0.99\columnwidth]{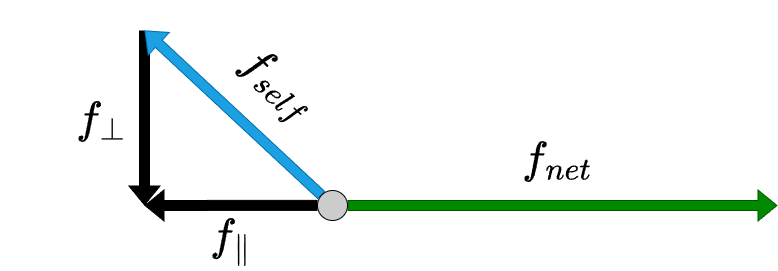}
    \caption{When $f_{self}$ of the agents is obtuse relative to the net force.}
    \label{fig:obtuse}
\end{figure}

\subsection{Measured Force is Obtuse to the Net Force}

In the obtuse angle ranges, $f_\perp$ decreases, and $f_\parallel$ becomes more negative as the angle between $f_{self}$ and the net force increases. This is demonstrated in Figure \ref{fig:obtuse}. Again, based on hypotheses in the literature that people are optimizing for minimal effort and highest efficiency, we would predict that the time spent in this region would be minimal for any given HH co-manipulation because by operating in these ranges participants begin to directly oppose the acceleration of the system. However, forces that are obtuse with respect to the net force occur 29.8$\%$ of the time in the LL study, 27.9$\%$ of the time in the LF study, and only 16.8$\%$ of the time in the HR study (see Table \ref{tab:Angle_table}). Contrary to expectations, the LL study spends more time in these ranges than other studies. This further supports the idea that participants are not aiming for minimal effort or highest efficiency, but that they have some other preferred state or hierarchy of objectives.

\begin{figure}[tb]
    \includegraphics[width=0.99\columnwidth]{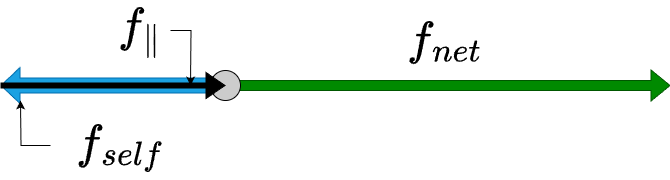}
    \caption{Antagonistic case, where the measured force is anti-aligned with the net force, not necessarily other agents. }
    \label{fig:antagonist}
\end{figure}

\subsection{Measured Force is Antagonistic to the Net Force }

Finally, the antagonist range is where all, or nearly all, of $f_{self}$ is acting against the net force as shown in Figure \ref{fig:antagonist}. In this condition, the agent force is 180 degrees, $\pm$ 5 degrees, from the net force and in direct opposition to the acceleration of the system. This is the worst-case scenario from the perspective of efficiency and cooperation. Under the same hypothesis as before, we would predict this case to occur rarely. This case occurs 1.45$\%$ of the time in the LL study, 1.37$\%$ of the time in the HR study, and 0.78$\%$ of the time in the HR study, as summarized in Table \ref{tab:Angle_table}. While this range is infrequent, it is surprisingly consistent across HH studies and remarkably low in the HR study. The fact that the HR study was in the antagonist and obtuse ranges for the least amount of time indicates that it was more efficient than the HH studies. Again we see that our ideal case with the LL study spent the most time in this range. This again raises the question: What do people prefer? It seems unlikely that the antagonistic range is the preferred state as it happens so much less frequently than either the aligned or orthogonal states. 

\begin{figure}[tb]
    \includegraphics[width=3.5 in]{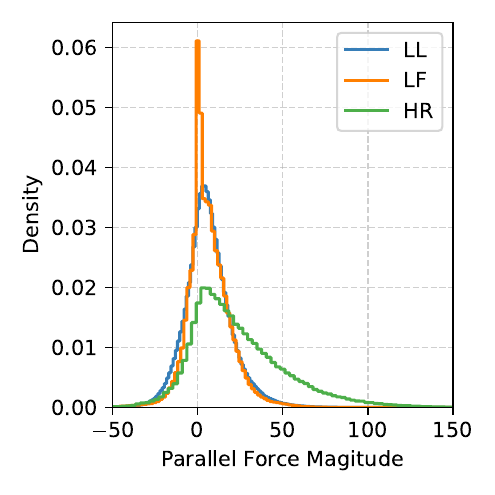} 
    \caption{Histogram showing the frequency and magnitude of $f_\parallel$ for each study. Not all data is shown due to a large range of outliers.}
    \label{fig:ll_hist}
\end{figure}

\begin{figure}[!htb]
    \includegraphics[width=3.5 in]{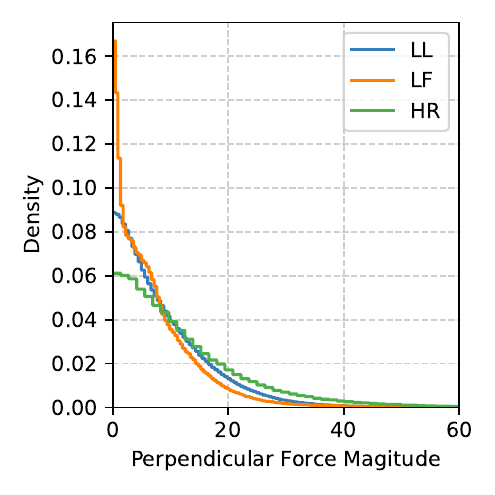} 
    \caption{Histogram showing the frequency and magnitude of $f_\perp$ for each study. Not all data is shown due to a large range of outliers.}
    \label{fig:perp_hist}
\end{figure}

Our new decomposition provides a consistent measure that allows for n agents, 4 DOFs, and independence of the choice of reference frame. It also captures significant differences between HH trials and current HR trials. Evidence for these differences can be seen when comparing $f_\parallel$ and $f_\perp$ forces across each of the three studies.  By creating a histogram of the magnitudes of each of these forces it can be clearly observed in Figures \ref{fig:ll_hist}, and \ref{fig:perp_hist} that the HR study has a heavy-tailed skew towards greater magnitudes of both $f_\parallel$ and $f_\perp$ forces. This provides a starting place to look for key differences when attempting to improve HR co-manipulation. In addition to this, the remarkable similarity between HH trials suggests a consistent pattern or method that participants are using, which may be exploited to produce better HR performance. Finally, note the large peak at zero in the LF study. This peak suggests that in the LF scenario, the follower defaults to doing nothing, applying very little if any force to the system, while being very similar to the LL study the rest of the time.

\section{Tension vs Compression}
\label{sec:TC}

Having disabused the hypothesis that participants default to, or strive for, the most efficient and minimal effort states from the perspective of wasted force, we propose a new hypothesis: participants prefer to keep the system in tension. Putting the table in a state of tension could serve two purposes. First, it puts more space between the participants and the table itself, which reduces the chances of colliding with the table and allows participants more time to react when the table is pushed towards them. To help test this hypothesis, a consistent measure of tension and compression was developed that meets our requirements. Given the vector location of $f_{self}$ relative to the center of mass, tension and compression force can be calculated via the following equation:

\begin{equation}
\left|\frac{f_{min} \cdot P_{min}}{||P_{min}||} \right|\left(\frac{f_0 \cdot P_0}{||f_0 \cdot P_0||}+\frac{(f_{net}-F_0) \cdot P_{net}}{||(f_{net}-f_0) \cdot P_{net}||}\right)\div2
\end{equation}

\begin{figure}[tb]    
    \includegraphics[width=0.99\columnwidth]{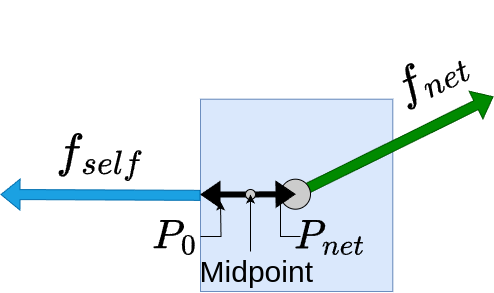}
    \caption{Free body diagram for the calculation of tension and compression.}
    \label{fig:TC_diagram}
\end{figure}

\begin{figure}[!htb]
    \includegraphics[width=3.5 in]{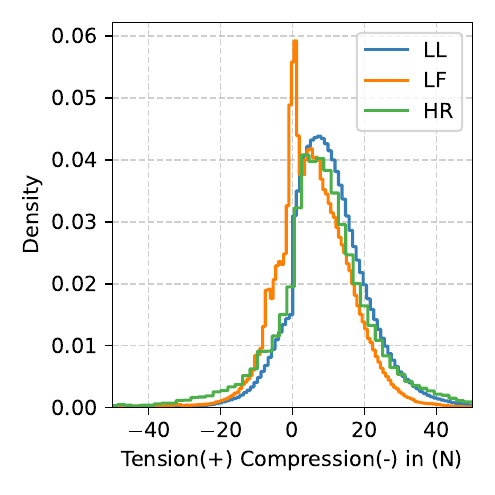} 
    \caption{Histograms of tension and compression for each of the three studies. Not all data is shown due to a large range of outliers.}
    \label{fig:TC_hist}
\end{figure}

Where $f_0$ = $f_{self}$ is the force measured at the input position of the agent of interest and $f_{net}$ is the net force or force accelerating the system. The midpoint is the center of the line segment connecting the location of the center of mass and the location of $f_{self}$. $P_0$ is a position vector from the midpoint to the position of $f_{self}$, and $P_{net}$ is a position vector from the midpoint to the center of mass. $f_{min}$ is the vector with the smallest magnitude between $f_{self}$ and $f_{net}$, and $P_{min}$ is the position of the smaller of the two forces $F_0$ and $f_{net}$. Positions are measured from the midpoint of the two input vectors as pictured in Figure \ref{fig:TC_diagram}. Positive values indicate tension and negative values indicate compression. The value itself indicates the magnitude of tension or compression between the agent and the other forces acting at the center of mass of the system. This measure was applied to the following three studies: \cite{freeman_motion_2024, mielke_analysis_2017, Mielke_HR_data_driven_2024}. We again observe the tendency towards doing nothing in the LF study, present as a sharp peak about zero as can be observed in Figure \ref{fig:TC_hist}. The LF study also contains a small bulge on the compression side of the histogram which is likely caused by not knowing exactly when to start and stop. 

The idea that tension might be a preferred or default state for participants in co-manipulation tasks, explored herein, is well supported. We offer the following three reasons. First, significantly more time is spent in tension than in either compression or cooperation as reported in Table \ref{tab:TC_table}. Second, the LL study spent more time in this state than any other study, see Table \ref{tab:TC_table}. The shared knowledge of the goal enabled the LL study to spend less time and effort communicating the goal haptically and more time in preferred states. Third, the histograms from all three studies have very similar distributions on the tension side, as can be seen in Figure \ref{fig:TC_hist}, suggesting that when a human is involved in co-manipulation, whether with a robot or another human, they tend to exhibit the same tensioning behaviors. 

A common point of confusion is how tension can be the preferred state when so little time is spent in the antagonistic range. This is because the antagonistic range is defined here with respect to the net force, not the other participant(s). This means that $f_{self}$ can be in any of the five ranges relative to the net force while the object is in a state of tension. This improved understanding of tension as a desired or default state will be useful when designing a control method for co-manipulation; we can now plan to leave some amount of tension without interpreting it as a signal to move. Additionally, actively maintaining the right amount of tension seems likely to help improve the human-robot co-manipulation experience, making robots more comfortable to work with.

\begin{table}[tb]
\centering
\caption{Time spent in tension, compression, and cooperation for three separate studies. Percentages are calculated by dividing the number of measures in each state by the total number of measures reported in each study. Measures were collected at 200$hz$ }
\label{tab:TC_table}
\begin{tabular}{lrrrr}\toprule
&Tension &Compression &Cooperation \\\midrule
LL &72.4\% &10.1\% &17.5\% \\
LF &67.6\% &24.6\% &7.8\% \\
HR &63.9\% &15.5\% &20.6\% \\
\bottomrule
\end{tabular}
\end{table}

\section{Conclusion}

This paper proposes the ``net force method'', to decompose interaction force. This method is independent of the number of agents and their locations and input wrenches, extends to four DOFs, can be obtained by a self-contained robotic system, is invariant with respect to a chosen frame of reference, and can be used for objects of significant mass. The net force method accomplishes this by decomposing a measured ($f_{self}$) force into two components that are parallel and perpendicular to the net force ($f_\parallel$ and $f_\perp$). This provides a consistent way of ascertaining the state of an agent relative to the state of the group. Viewing co-manipulation data from the perspective of the angle between $f_{self}$ and the net force, and from the perspective of the $f_\parallel$ and $f_\perp$ components produced by the net force method, provides key insights into human behaviors and how they differ from current state of the art human-robot co-manipulation methods. Future work should include a more sophisticated gravity compensation method to extend the measure to a complete six DOFs. Further work should also be done to develop a mapping from $f_\parallel$ and $f_\perp$ measures to specific actions or behaviors for use in robotic control. This base-level behavior will be highly beneficial to accomplish before seeking to implement higher-level control such as path planning and obstacle avoidance.

\section*{Acknowledgment}

The authors would like to thank the National Science Foundation’s  National Robotics Initiative (NRI) under grant number 2024792 for their generous funding that made this research possible.

\bibliography{main}

\begin{thebibliography}{10}
\providecommand{\url}[1]{#1}
\csname url@samestyle\endcsname
\providecommand{\newblock}{\relax}
\providecommand{\bibinfo}[2]{#2}
\providecommand{\BIBentrySTDinterwordspacing}{\spaceskip=0pt\relax}
\providecommand{\BIBentryALTinterwordstretchfactor}{4}
\providecommand{\BIBentryALTinterwordspacing}{\spaceskip=\fontdimen2\font plus
\BIBentryALTinterwordstretchfactor\fontdimen3\font minus \fontdimen4\font\relax}
\providecommand{\BIBforeignlanguage}[2]{{%
\expandafter\ifx\csname l@#1\endcsname\relax
\typeout{** WARNING: IEEEtran.bst: No hyphenation pattern has been}%
\typeout{** loaded for the language `#1'. Using the pattern for}%
\typeout{** the default language instead.}%
\else
\language=\csname l@#1\endcsname
\fi
#2}}
\providecommand{\BIBdecl}{\relax}
\BIBdecl

\bibitem{takagi_individuals_2019}
\BIBentryALTinterwordspacing
A.~Takagi, M.~Hirashima, D.~Nozaki, and E.~Burdet, ``Individuals physically interacting in a group rapidly coordinate their movement by estimating the collective goal,'' \emph{eLife}, vol.~8, p. e41328, Feb. 2019, publisher: eLife Sciences Publications, Ltd. [Online]. Available: \url{https://doi.org/10.7554/eLife.41328}
\BIBentrySTDinterwordspacing

\bibitem{reed_kinesthetic_2005}
\BIBentryALTinterwordspacing
K.~Reed, M.~Peshkin, J.~Colgate, and J.~Patton, ``Kinesthetic {Interaction},'' \emph{9th International Conference on Rehabilitation Robotics, 2005}, pp. 569--574, Jan. 2005. [Online]. Available: \url{https://digitalcommons.usf.edu/egr_facpub/123}
\BIBentrySTDinterwordspacing

\bibitem{bussy_proactive_2012}
A.~Bussy, P.~Gergondet, A.~Kheddar, F.~Keith, and A.~Crosnier, ``Proactive behavior of a humanoid robot in a haptic transportation task with a human partner,'' in \emph{2012 {IEEE} {RO}-{MAN}: {The} 21st {IEEE} {International} {Symposium} on {Robot} and {Human} {Interactive} {Communication}}, Sep. 2012, pp. 962--967, iSSN: 1944-9437.

\bibitem{erhart_internal_2015}
\BIBentryALTinterwordspacing
S.~Erhart and S.~Hirche, ``Internal {Force} {Analysis} and {Load} {Distribution} for {Cooperative} {Multi}-{Robot} {Manipulation},'' \emph{IEEE Transactions on Robotics}, vol.~31, no.~5, pp. 1238--1243, Oct. 2015, conference Name: IEEE Transactions on Robotics. [Online]. Available: \url{https://ieeexplore.ieee.org/document/7206596}
\BIBentrySTDinterwordspacing

\bibitem{schmidts_new_2016}
\BIBentryALTinterwordspacing
A.~M. Schmidts, M.~Schneider, M.~Kühne, and A.~Peer, ``A new interaction force decomposition maximizing compensating forces under physical work constraints,'' in \emph{2016 {IEEE} {International} {Conference} on {Robotics} and {Automation} ({ICRA})}, May 2016, pp. 4922--4929. [Online]. Available: \url{https://ieeexplore.ieee.org/document/7487698}
\BIBentrySTDinterwordspacing

\bibitem{sawers_small_2017}
\BIBentryALTinterwordspacing
A.~Sawers, T.~Bhattacharjee, J.~L. McKay, M.~E. Hackney, C.~C. Kemp, and L.~H. Ting, ``\BIBforeignlanguage{en}{Small forces that differ with prior motor experience can communicate movement goals during human-human physical interaction},'' \emph{\BIBforeignlanguage{en}{Journal of NeuroEngineering and Rehabilitation}}, vol.~14, no.~1, p.~8, Jan. 2017. [Online]. Available: \url{https://doi.org/10.1186/s12984-017-0217-2}
\BIBentrySTDinterwordspacing

\bibitem{lanini_interactive_2017}
\BIBentryALTinterwordspacing
J.~Lanini, A.~Duburcq, H.~Razavi, C.~G.~L. Goff, and A.~J. Ijspeert, ``\BIBforeignlanguage{en}{Interactive locomotion: {Investigation} and modeling of physically-paired humans while walking},'' \emph{\BIBforeignlanguage{en}{PLOS ONE}}, vol.~12, no.~9, p. e0179989, Sep. 2017, publisher: Public Library of Science. [Online]. Available: \url{https://journals.plos.org/plosone/article?id=10.1371/journal.pone.0179989}
\BIBentrySTDinterwordspacing

\bibitem{kumar_force_1988}
\BIBentryALTinterwordspacing
V.~Kumar and K.~Waldron, ``Force distribution in closed kinematic chains,'' \emph{IEEE Journal on Robotics and Automation}, vol.~4, no.~6, pp. 657--664, Dec. 1988, conference Name: IEEE Journal on Robotics and Automation. [Online]. Available: \url{https://ieeexplore.ieee.org/abstract/document/9303}
\BIBentrySTDinterwordspacing

\bibitem{groten_experimental_2009}
R.~Groten, D.~Feth, H.~Goshy, A.~Peer, D.~A. Kenny, and M.~Buss, ``Experimental analysis of dominance in haptic collaboration,'' in \emph{{RO}-{MAN} 2009 - {The} 18th {IEEE} {International} {Symposium} on {Robot} and {Human} {Interactive} {Communication}}, Sep. 2009, pp. 723--729, iSSN: 1944-9437.

\bibitem{mortl_role_2012}
\BIBentryALTinterwordspacing
A.~Mörtl, M.~Lawitzky, A.~Kucukyilmaz, M.~Sezgin, C.~Basdogan, and S.~Hirche, ``\BIBforeignlanguage{en}{The role of roles: {Physical} cooperation between humans and robots},'' \emph{\BIBforeignlanguage{en}{The International Journal of Robotics Research}}, vol.~31, no.~13, pp. 1656--1674, Nov. 2012, publisher: SAGE Publications Ltd STM. [Online]. Available: \url{https://doi.org/10.1177/0278364912455366}
\BIBentrySTDinterwordspacing

\bibitem{noohi_model_2016}
E.~Noohi, M.~Žefran, and J.~L. Patton, ``A {Model} for {Human}–{Human} {Collaborative} {Object} {Manipulation} and {Its} {Application} to {Human}–{Robot} {Interaction},'' \emph{IEEE Transactions on Robotics}, vol.~32, no.~4, pp. 880--896, Aug. 2016, conference Name: IEEE Transactions on Robotics.

\bibitem{mielke_human-robot_2020}
\BIBentryALTinterwordspacing
E.~Mielke, E.~Townsend, D.~Wingate, and M.~D. Killpack, ``\BIBforeignlanguage{en}{Human-robot co-manipulation of extended objects: {Data}-driven models and control from analysis of human-human dyads},'' \emph{\BIBforeignlanguage{en}{arXiv:2001.00991 [cs, eess]}}, Jan. 2020, arXiv: 2001.00991. [Online]. Available: \url{http://arxiv.org/abs/2001.00991}
\BIBentrySTDinterwordspacing

\bibitem{jensen_trends_2021}
S.~W. Jensen, J.~L. Salmon, and M.~D. Killpack, ``\BIBforeignlanguage{eng}{Trends in {Haptic} {Communication} of {Human}-{Human} {Dyads}: {Toward} {Natural} {Human}-{Robot} {Co}-manipulation},'' \emph{\BIBforeignlanguage{eng}{Frontiers in Neurorobotics}}, vol.~15, p. 626074, 2021.

\bibitem{jlassi_online_2014}
S.~Jlassi, S.~Tliba, and Y.~Chitour, ``An {Online} {Trajectory} generator-{Based} {Impedance} control for co-manipulation tasks,'' in \emph{2014 {IEEE} {Haptics} {Symposium} ({HAPTICS})}, Feb. 2014, pp. 391--396, iSSN: 2324-7355.

\bibitem{cremer_robotic_2014}
S.~Cremer, I.~Ranatunga, and D.~O. Popa, ``Robotic waiter with physical co-manipulation capabilities,'' in \emph{2014 {IEEE} {International} {Conference} on {Automation} {Science} and {Engineering} ({CASE})}, Aug. 2014, pp. 1153--1158, iSSN: 2161-8089.

\bibitem{Mielke_HR_data_driven_2024}
\BIBentryALTinterwordspacing
E.~Mielke, E.~Townsend, D.~Wingate, J.~L. Salmon, and M.~D. Killpack, ``Human-robot planar co-manipulation of extended objects: data-driven models and control from human-human dyads,'' \emph{Frontiers in Neurorobotics}, vol.~18, 2024. [Online]. Available: \url{https://www.frontiersin.org/journals/neurorobotics/articles/10.3389/fnbot.2024.1291694}
\BIBentrySTDinterwordspacing

\bibitem{donner_physically_2018}
\BIBentryALTinterwordspacing
P.~Donner, S.~Endo, and M.~Buss, ``Physically {Plausible} {Wrench} {Decomposition} for {Multieffector} {Object} {Manipulation},'' \emph{IEEE Transactions on Robotics}, vol.~34, no.~4, pp. 1053--1067, Aug. 2018, conference Name: IEEE Transactions on Robotics. [Online]. Available: \url{https://ieeexplore.ieee.org/document/8375105}
\BIBentrySTDinterwordspacing

\bibitem{karayiannidis_online_2013}
Y.~Karayiannidis, C.~Smith, F.~E. Viña, and D.~Kragic, ``Online kinematics estimation for active human-robot manipulation of jointly held objects,'' in \emph{2013 {IEEE}/{RSJ} {International} {Conference} on {Intelligent} {Robots} and {Systems}}, Nov. 2013, pp. 4872--4878, iSSN: 2153-0866.

\bibitem{duchaine_general_2007}
V.~Duchaine and C.~M. Gosselin, ``General {Model} of {Human}-{Robot} {Cooperation} {Using} a {Novel} {Velocity} {Based} {Variable} {Impedance} {Control},'' in \emph{Second {Joint} {EuroHaptics} {Conference} and {Symposium} on {Haptic} {Interfaces} for {Virtual} {Environment} and {Teleoperator} {Systems} ({WHC}'07)}, Mar. 2007, pp. 446--451.

\bibitem{ikeura_cooperative_1995}
R.~Ikeura and H.~Inooka, ``Cooperative force control in carrying an object by two humans,'' in \emph{1995 {IEEE} {International} {Conference} on {Systems}, {Man} and {Cybernetics}. {Intelligent} {Systems} for the 21st {Century}}, vol.~3, Oct. 1995, pp. 2307--2311 vol.3.

\bibitem{freeman_motion_2024}
\BIBentryALTinterwordspacing
S.~Freeman, S.~Moss, J.~L. Salmon, and M.~D. Killpack, ``Classification of co-manipulation modus with human-human teams for future application to human-robot systems,'' \emph{J. Hum.-Robot Interact.}, jun 2024, just Accepted. [Online]. Available: \url{https://doi.org/10.1145/3659059}
\BIBentrySTDinterwordspacing

\bibitem{parker_experimental_2011}
C.~A.~C. Parker and E.~A. Croft, ``Experimental investigation of human-robot cooperative carrying,'' in \emph{2011 {IEEE}/{RSJ} {International} {Conference} on {Intelligent} {Robots} and {Systems}}, Sep. 2011, pp. 3361--3366, iSSN: 2153-0866.

\bibitem{corteville_human-inspired_2007}
B.~Corteville, E.~Aertbelien, H.~Bruyninckx, J.~De~Schutter, and H.~Van~Brussel, ``Human-inspired robot assistant for fast point-to-point movements,'' in \emph{Proceedings 2007 {IEEE} {International} {Conference} on {Robotics} and {Automation}}, Apr. 2007, pp. 3639--3644, iSSN: 1050-4729.

\bibitem{melendez-calderon_interpersonal_2015}
\BIBentryALTinterwordspacing
A.~Melendez-Calderon, V.~Komisar, and E.~Burdet, ``Interpersonal strategies for disturbance attenuation during a rhythmic joint motor action,'' \emph{Physiology \& Behavior}, vol. 147, pp. 348--358, Aug. 2015. [Online]. Available: \url{https://www.sciencedirect.com/science/article/pii/S003193841500253X}
\BIBentrySTDinterwordspacing

\bibitem{takagi_motion_2016}
\BIBentryALTinterwordspacing
A.~Takagi, N.~Beckers, and E.~Burdet, ``\BIBforeignlanguage{en}{Motion {Plan} {Changes} {Predictably} in {Dyadic} {Reaching}},'' \emph{\BIBforeignlanguage{en}{PLOS ONE}}, vol.~11, no.~12, p. e0167314, Dec. 2016, publisher: Public Library of Science. [Online]. Available: \url{https://journals.plos.org/plosone/article?id=10.1371/journal.pone.0167314}
\BIBentrySTDinterwordspacing

\bibitem{takagi_physically_2017}
\BIBentryALTinterwordspacing
A.~Takagi, G.~Ganesh, T.~Yoshioka, M.~Kawato, and E.~Burdet, ``\BIBforeignlanguage{en}{Physically interacting individuals estimate the partner’s goal to enhance their movements},'' \emph{\BIBforeignlanguage{en}{Nature Human Behaviour}}, vol.~1, no.~3, pp. 1--6, Mar. 2017, publisher: Nature Publishing Group. [Online]. Available: \url{https://www.nature.com/articles/s41562-017-0054}
\BIBentrySTDinterwordspacing

\bibitem{takagi_haptic_2018}
\BIBentryALTinterwordspacing
A.~Takagi, F.~Usai, G.~Ganesh, V.~Sanguineti, and E.~Burdet, ``Haptic communication between humans is tuned by the hard or soft mechanics of interaction,'' \emph{PLoS Computational Biology}, vol.~14, no.~3, p. e1005971, Mar. 2018. [Online]. Available: \url{https://www.ncbi.nlm.nih.gov/pmc/articles/PMC5863953/}
\BIBentrySTDinterwordspacing

\bibitem{takagi_flexible_2021}
A.~Takagi, Y.~Li, and E.~Burdet, ``Flexible {Assimilation} of {Human}'s {Target} for {Versatile} {Human}-{Robot} {Physical} {Interaction},'' \emph{IEEE Transactions on Haptics}, vol.~14, no.~2, pp. 421--431, Apr. 2021, conference Name: IEEE Transactions on Haptics.

\bibitem{peternel_towards_2016}
L.~Peternel, N.~Tsagarakis, and A.~Ajoudani, ``Towards multi-modal intention interfaces for human-robot co-manipulation,'' in \emph{2016 {IEEE}/{RSJ} {International} {Conference} on {Intelligent} {Robots} and {Systems} ({IROS})}, Oct. 2016, pp. 2663--2669, iSSN: 2153-0866.

\bibitem{reed_haptic_2006}
K.~B. Reed, M.~Peshkin, M.~J. Hartmann, J.~Patton, P.~M. Vishton, and M.~Grabowecky, ``Haptic cooperation between people, and between people and machines,'' in \emph{2006 {IEEE}/{RSJ} {International} {Conference} on {Intelligent} {Robots} and {Systems}}, Oct. 2006, pp. 2109--2114, iSSN: 2153-0866.

\bibitem{reed_physical_2008}
K.~B. Reed and M.~A. Peshkin, ``Physical {Collaboration} of {Human}-{Human} and {Human}-{Robot} {Teams},'' \emph{IEEE Transactions on Haptics}, vol.~1, no.~2, pp. 108--120, Jul. 2008, conference Name: IEEE Transactions on Haptics.

\bibitem{mielke_analysis_2017}
\BIBentryALTinterwordspacing
E.~Mielke, E.~Townsend, and M.~Killpack, ``\BIBforeignlanguage{en}{Analysis of {Rigid} {Extended} {Object} {Co}-{Manipulation} by {Human} {Dyads}: {Lateral} {Movement} {Characterization}},'' in \emph{\BIBforeignlanguage{en}{Robotics: {Science} and {Systems} {XIII}}}.\hskip 1em plus 0.5em minus 0.4em\relax Robotics: Science and Systems Foundation, Jul. 2017. [Online]. Available: \url{http://www.roboticsproceedings.org/rss13/p47.pdf}
\BIBentrySTDinterwordspacing

\bibitem{madan_recognition_2015}
C.~E. Madan, A.~Kucukyilmaz, T.~M. Sezgin, and C.~Basdogan, ``Recognition of {Haptic} {Interaction} {Patterns} in {Dyadic} {Joint} {Object} {Manipulation},'' \emph{IEEE Transactions on Haptics}, vol.~8, no.~1, pp. 54--66, Jan. 2015, conference Name: IEEE Transactions on Haptics.

\bibitem{bing-ran_zuo_equivalence_1999}
\BIBentryALTinterwordspacing
{Bing-Ran Zuo} and {Wen-Han Qian}, ``\BIBforeignlanguage{en}{On the equivalence of internal and interaction forces in multifingered grasping},'' \emph{\BIBforeignlanguage{en}{IEEE Transactions on Robotics and Automation}}, vol.~15, no.~5, pp. 934--941, Oct. 1999. [Online]. Available: \url{http://ieeexplore.ieee.org/document/795796/}
\BIBentrySTDinterwordspacing

\bibitem{townsend_estimating_2017}
\BIBentryALTinterwordspacing
E.~C. Townsend, E.~A. Mielke, D.~Wingate, and M.~D. Killpack, ``Estimating {Human} {Intent} for {Physical} {Human}-{Robot} {Co}-{Manipulation},'' \emph{arXiv:1705.10851 [cs]}, May 2017, arXiv: 1705.10851. [Online]. Available: \url{http://arxiv.org/abs/1705.10851}
\BIBentrySTDinterwordspacing

\bibitem{allin_measuring_2002}
S.~Allin, Y.~Matsuoka, and R.~Klatzky, ``Measuring just noticeable differences for haptic force feedback: implications for rehabilitation,'' in \emph{Proceedings 10th {Symposium} on {Haptic} {Interfaces} for {Virtual} {Environment} and {Teleoperator} {Systems}. {HAPTICS} 2002}, Mar. 2002, pp. 299--302.

\end{thebibliography}
\bibliographystyle{IEEEtran}

\appendices

\section{H-H and H-R Experiment Descriptions}
\label{sec:appendix}

\subsection{Human-human Co-manipulation Study on Visual Input}

\begin{figure}[htb!]
    \centering
    \includegraphics[width=3.5 in]{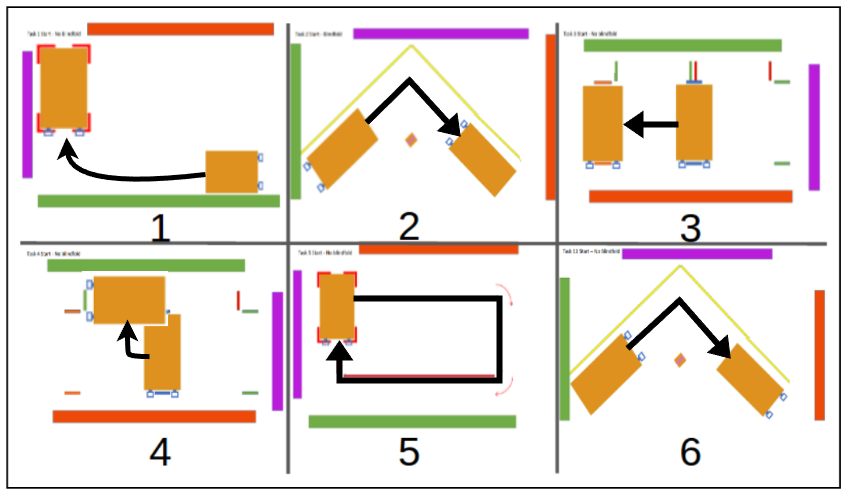} 
    \caption{Tasks performed by participants in Mielke et al.'s \cite{mielke_analysis_2017} study. Starting positions (top) and ending positions (bottom), figures are used with permission. }
    \label{fig:LF_tasks}
\end{figure}

\begin{figure}[htb!]
    \centering
    \includegraphics[width=3.5 in]{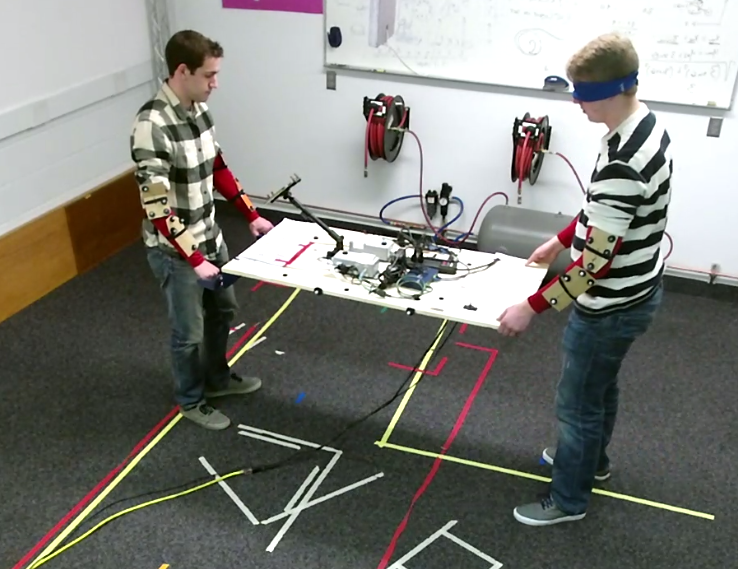} 
    \caption{Setup used in Mielke et al.'s extended co-manipulation experiment, image used with permission. \cite{mielke_analysis_2017}.}
    \label{fig:box_m}
\end{figure}

The data from \cite{mielke_analysis_2017} was collected from 21 human-human (HH) dyads consisting of 26 males and 16 females. In each dyad, one participant was randomly selected to be the leader and was given the instructions for each task. The other was designated as a follower, who was not given any task-related instructions and was also given a blindfold to be used randomly on half of the tasks throughout the experiment. Each dyad completed six tasks, as shown in Figure ~\ref{fig:LF_tasks}, a total of six times each. Data was collected by the object shown in Figure ~\ref{fig:box_m}. An IR motion capture system tracked the object and arm positions of the dyad and forces were tracked via a pair of force/torque sensors on the leader side. All other measures were calculated from position, time, and force data. A brief video of the study is available at \href{https://youtu.be/i-s1pIs17oY}{https://youtu.be/i-s1pIs17oY}.

\subsection{Human-Robot Co-Manipulation Study on Control Methods}

\begin{figure}[htb!]
    \centering
    \includegraphics[width=1\linewidth]{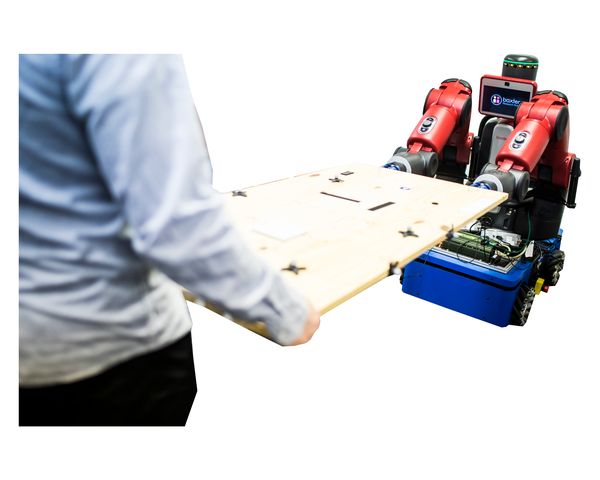} 
    \caption{Setup used in Mielke et al.'s human-robot experiment in an extended co-manipulation experiment, image used with permission \cite{Mielke_HR_data_driven_2024}.}
    \label{fig:HR_setup}
\end{figure}

\begin{figure}[htb!]
    \centering
    \includegraphics[width=1\linewidth]{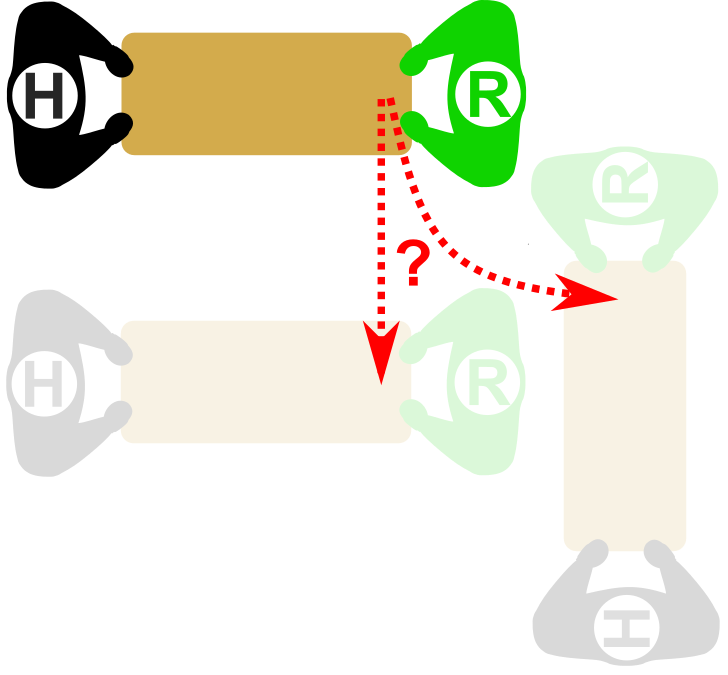} 
    \caption{Tasks performed by participants in Mielke et al.'s human-robot study. Where H stands for human, R stands for robot, and the dotted arrows refer to the translation vs rotation problem \cite{Mielke_HR_data_driven_2024}. Figure used with permission. }
    \label{fig:HR_tasks}
\end{figure}

Mielke et al. conducted a follow-up study with a human-robot dyad, as shown in Figure \ref{fig:HR_setup} (see \cite{Mielke_HR_data_driven_2024}). This study included a set of 16 human participants performing a set of two tasks with a robot partner. These tasks consisted of a translation or rotation, as shown in Figure \ref{fig:HR_tasks}, and were completed eight times with two different control algorithms. These tasks were performed to test a solution to the translation vs rotation problem, described by Mielke as follows: ``When an extended object is included in co-manipulation tasks, forces applied in a lateral direction could indicate either intent to translate laterally, or intent to rotate the object in the plane" \cite{Mielke_HR_data_driven_2024}. A brief video of the study is available at \href{https://youtu.be/QQKpT1ORxkw}{https://youtu.be/QQKpT1ORxkw}.

\subsection{Human-Human Co-Manipulation Study on Modus}
\begin{figure}[tb]
    \centering
    \includegraphics[width=1\linewidth]{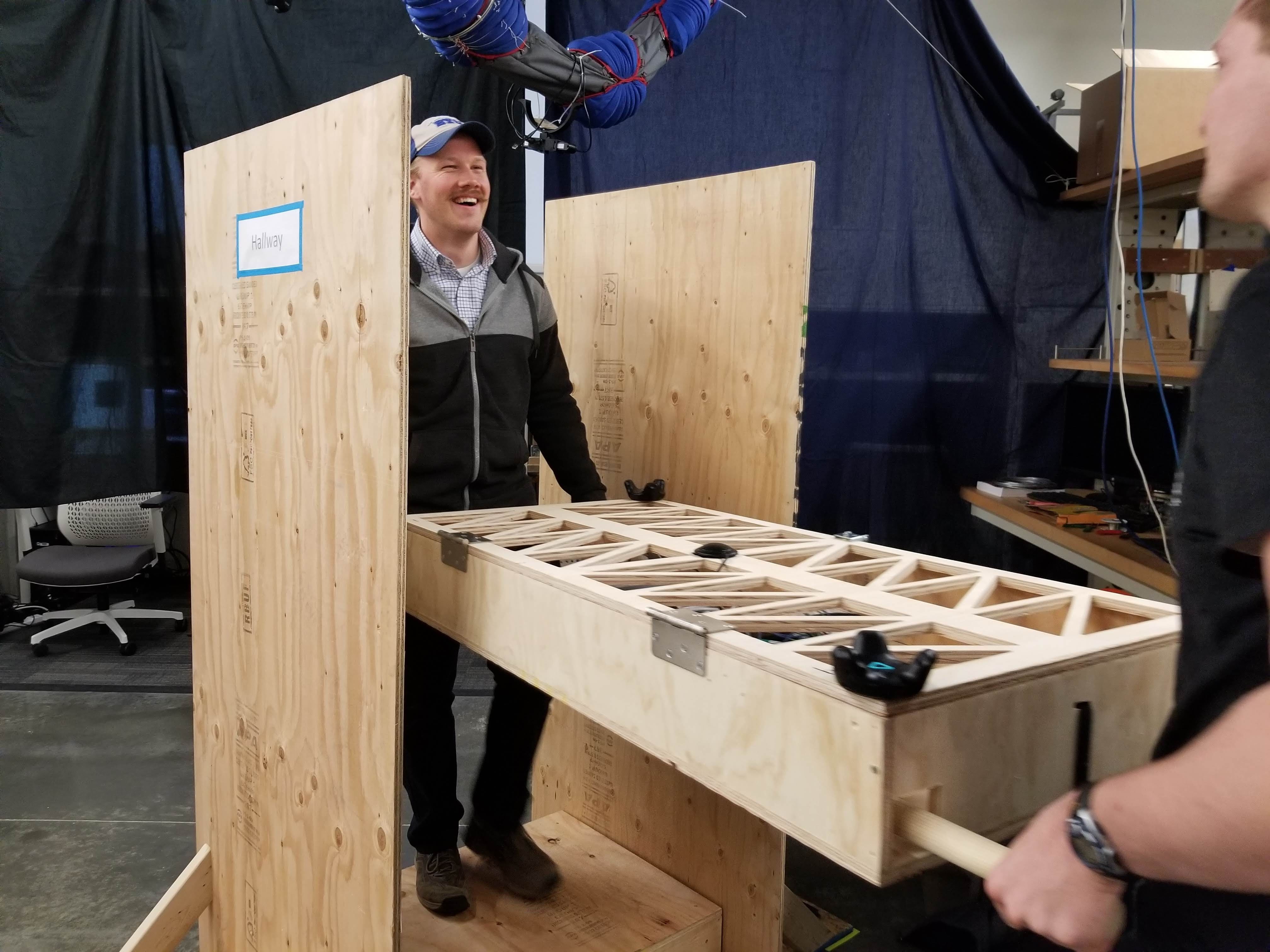} 
    \caption{Setup used in Freeman's extended co-manipulation experiment, image used with permission. \cite{freeman_motion_2024}}
    \label{fig:LL_setup}
\end{figure}

\begin{figure}[tb!]
    \centering
    \includegraphics[width=1\linewidth]{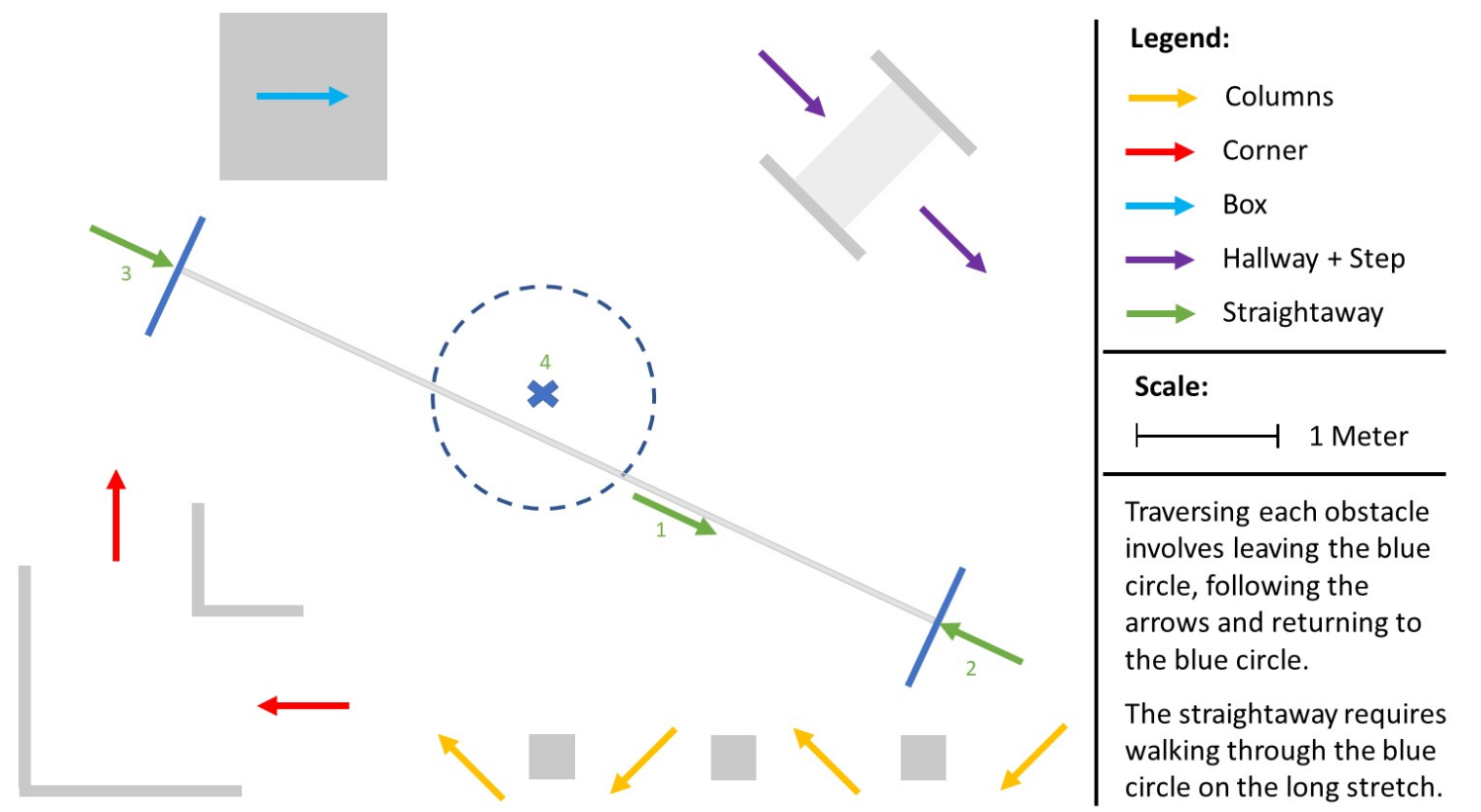} 
    \caption{Tasks performed in Freeman's extended co-manipulation experiment, figure used with permission. \cite{freeman_motion_2024}}
    \label{fig:obs}
\end{figure}

Freeman \cite{freeman_motion_2024} conducted an experiment to determine if the intent or desired modus, such as trying to move cautiously, quickly, or smoothly could be determined from haptic information. In other words, given haptic information, could a robot determine if its human partner wanted to move quickly, smoothly, etc? Freeman’s data was collected from 19 male dyads and 11 female dyads who were directed through a series of five obstacles, shown in Figure ~\ref{fig:obs}, carrying an object, see Figure ~\ref{fig:LL_setup} with force/torque sensors on both ends. In a randomized order, dyads were given six different modi (i.e. quickly, smoothly, obstacle avoidance, no talking, no context, and tutorial), defined using realistic scenarios to motivate the dyads to behave accordingly. A brief video of the study is available \href{https://youtu.be/eg2mf6LeDlo}{https://youtu.be/eg2mf6LeDlo}.

\end{document}